\documentclass[conference]{IEEEtran}

\usepackage{cite}
\usepackage{subfigure}
\usepackage{amsmath,amssymb,amsfonts}
\usepackage{algorithmic}
\usepackage{graphicx}
\usepackage{textcomp}
\usepackage{url}
\usepackage{xcolor}
\def\BibTeX{{\rm B\kern-.05em{\sc i\kern-.025em b}\kern-.08em
    T\kern-.1667em\lower.7ex\hbox{E}\kern-.125emX}}

\title{Evolving the Hearthstone Meta\\
}

\begin{document}

\author{\IEEEauthorblockN{Fernando de Mesentier Silva}
\IEEEauthorblockA{
\textit{Independent Researcher}\\
Rio de Janeiro, Brazil \\
fms2005@gmail.com}
\\
\\
\IEEEauthorblockN{Matthew C. Fontaine}
\IEEEauthorblockA{\textit{Independent Researcher} \\
Vermont, USA \\
tehqin@gmail.com}
\and
\IEEEauthorblockN{Rodrigo Canaan}
\IEEEauthorblockA{\textit{Game Innovation Lab} \\
\textit{New York University}\\
New York, USA \\
rodrigo.canaan@nyu.edu}
\\
\IEEEauthorblockN{Julian Togelius}
\IEEEauthorblockA{\textit{Game Innovation Lab} \\
\textit{New York University}\\
New York, USA \\
julian@togelius.com}
\and
\IEEEauthorblockN{Scott Lee}
\IEEEauthorblockA{
\textit{Independent Researcher}\\
California, USA \\
randomperson2727@gmail.com}
\\
\\
\IEEEauthorblockN{Amy K. Hoover}
\IEEEauthorblockA{\textit{Ying Wu College of Computing} \\
\textit{New Jersey Institute of Technology}\\
New Jersey, USA \\
ahoover@njit.edu}
}

\maketitle

\IEEEpubidadjcol

\begin{abstract}
Balancing an ever growing strategic game of high complexity, such as Hearthstone is a complex task. The target of making strategies diverse and customizable results in a delicate intricate system. Tuning over 2000 cards to generate the desired outcome without disrupting the existing environment becomes a laborious challenge. In this paper, we discuss the impacts that changes to existing cards can have on strategy in Hearthstone. By analyzing the win rate on match-ups across different decks, being played by different strategies, we propose to compare their performance before and after changes are made to improve or worsen different cards. Then, using an evolutionary algorithm, we search for a combination of changes to the card attributes that cause the decks to approach equal, 50\% win rates. We then expand our evolutionary algorithm to a multi-objective solution to search for this result, while making the minimum amount of changes, and as a consequence disruption, to the existing cards. Lastly, we propose and evaluate metrics to serve as heuristics with which to decide which cards to target with balance changes.
\end{abstract}


\section{Introduction}

Games are often complex systems with a myriad of moving parts. In such games, designers provide a large selection of game objects, such as cards, skills or equipment, allowing players to play the game in many different ways with varying combinations of these game objects. However in competitive games where players are motivated to win, players often focus on gathering particular combinations of objects that maximize their perceived chances of winning. Balancing these game objects is one of the most important tasks for game designers. In an unbalanced game, there may be a single degenerate strategy that is unbeatable or it may be impossible to play a particular character class well. Such a game is essentially broken and experienced players tend to quickly lose interest. 

Published by Blizzard, \emph{Hearthstone} is a popular collectible card game released in 2014 that contains over 2000 collectible cards. During a game, two players take turns trying to reduce the health of the opposing hero to zero. For any given game, a player chooses 30 cards to build a deck, and only these cards are accessible during gameplay. Cards from the deck are randomly drawn at each turn. Naturally, players try to pick combinations of cards that they think have the best chance of beating an opponent. While Blizzard sometimes edits existing cards (i.e.\ through \emph{nerfs} and \emph{buffs}), the longevity and popularity of the game indicates that the company is invested in keeping it well-tuned. 

In part because there are millions of subscribed players for Hearthstone, when new cards are released or existing cards edited, these parts of the strategy space are quickly explored by the players. Even a single misbalanced card could result in large disruptions to the \emph{metagame} (i.e.\ popular decks and their win rates against each other). A balanced metagame promotes a variety of competitive decks. While Blizzard balances the game through a combination of deep domain knowledge, in-house playtesting, and by examining the metagame as it develops through data visualization, in a game with over 2000 cards, there is still a need for computational tools to help. The question this paper addresses is how AI methods can facilitate game balancing in a game as complex as Hearthstone. 


This paper explores several methods for computationally balancing a collectible card game. While the focus here is on Hearthstone, the idea is that it is an exemplar for collectible card games. Methods in this paper could potentially apply to other games where players must choose from a wide collection of game objects to customize their characters.

The approach to game balancing in this paper is three-pronged. The first approach is based on the simplified assumption that decks should perform approximately equally against each other.  For this first experiment, we encode a set of card changes as an individual in an evolutionary algorithm, where good changes are those that equalize win rates between decks. This is a single objective optimization problem, where the fitness of individuals depends on how close the win rates for each deck matchup deviate from 50\%.


While after launch, some changes to a complex game are necessary to achieve balance, even small changes to cards can have a big effect on the metagame. Furthermore, such changes often require players to rethink their previous strategies, and in the case of Hearthstone they may need access to cards that they have yet to collect. Experiment 2 is a multi-objective optimization problem based on the same idea of balance as the first (i.e.\ decks achieve balance when they reach a 50\% win rate). However, one additional objective is to minimize changes.

While Experiment 1 and Experiment 2 holistically evolve changes for the entire set of cards, Experiment 3 explores the impact of targeting specific cards to balance. Focusing on a single deck, we first rank cards by how often the deck won if the player drew or played that card. Then we examine the impact of nerfing each card individually on the deck's win rates. We observe that nerfing cards that originally had a higher win rate when drawn (and to a lesser extent, win rate when played) tends to impact the deck's win rate more, suggesting that this could be a reasonable metric to guide  the search for effective balance changes in future experiments.


\section{Related Work}

Relevant background is described in this section including the game Hearthstone, previous approaches using AI in Hearthstone, and previous approaches of AI-techniques to balance games.

\subsection{Hearthstone}

Hearthstone is a Collectible Card Game (CCG) by Blizzard where players play as a hero from one of nine classes battling against each other to reduce their opponent's health points (HP) from a starting HP of 30 to 0. Before joining a match, players select a class and  build a deck of exactly 30 cards from a pool of class-specific cards plus a pool of neutral cards, which are available to any class.  Most cards cost a certain amount of mana, the game's main resource, to play. The amount of mana available on a player's turn starts at 1, and automatically increases by 1 per turn, to a maximum of 10.

Each card can be of one of four types. \emph{Minion} cards are player-controlled characters that can attack other minions or the enemy hero, dealing damage equal to their attack attribute. They are destroyed when dealt damage equal to or greater than their health attribute. \emph{Spell} cards are single-use effects applied when played, such as drawing cards, increasing health, summoning minions, dealing damage. \emph{Weapons} are a card type that a hero can equip to attack other minions or the opposing hero directly with, rather than through minion or spells. The last card type is a \emph{Hero} card, introduced in the sixth expansion. The hero cards replace a player's hero, sometimes changing the hero's properties. 

Additionally, choosing a class grants the player the ability to use that Hero's power, which is a small effect (such as dealing 2 damage to the enemy hero or losing 2 life in order to draw a card) that can be used once per turn by paying 2 mana, but without requiring a card in hand.

Decks can be loosely characterized by their favored strategy. The first common strategy this paper is concerned with is  ``Aggro", which attempts to win the game quickly, typically with a combination of cheap Minions and direct damage Spells, without trying to kill many opposing Minions. ``Control" attempts to win the game later, after controlling the board by using removal Spells and defensive Minions. Other strategies exist, such as ``Combo", which attempt to assemble a combination of highly synergistic cards which provide a large advantage or even win the game when played together. This paper focuses on the Aggro and Control strategies.

\subsection{AI and Hearthstone}

Being a popular and complex game, Hearthstone has been the subject of study by multiple researchers. While many efforts are focused on building gameplaying AIs~\cite{da2018hearthbot, dockhorn2018predicting,grad2017helping, santos2017monte,stiegler2017symbolic,swiechowski2018improving,zhang2017improving}, this subject is tangential to our work. To play the game, we use a breadth-first search together with a gamestate evaluating heuristic to approximate common player strategies such as Aggro or Control playstyles. This agent is part of the SabberStone~\footnote{SabberStone - https://github.com/HearthSim/SabberStone} engine we use to emulate the game. While sub optimal, the agent performance is sufficient in terms of quality and time for our experiments.

Evolutionary algorithms have been employed in Hearthstone primarily to search the deck space~\cite{bhatt2018exploring, fontaine:gecco19,garcia2016evolutionary,garcia2018automated}. The problem of creating a deck is parallel to that of playing the game. Finding a high performance card set is a non-trivial task that is beyond the scope of this work. Rather, we utilize the decks found by the quality diversity algorithm Map-Elites~\cite{mouret2015illuminating} when searching the space of cards from the Basic and Classic sets of Hearthstone~\cite{fontaine:gecco19}. Furthermore, deck building in Hearthstone has been investigated outside the scope of evolutionary algorithms~\cite{chen2018q,stiegler2016hearthstone}, and evolutionary algorithms have been used to create decks for other games, such as Magic: The Gathering~\cite{bjorke2017deckbuilding}.

Win rate prediction for Hearthstone decks has been explored~\cite{jakubik2017evaluation}, specially with the introduction of the AAAI Data Mining Competition~\cite{janusz2018toward}. Most approaches are not applicable to the problem we are introducing, considering they rely on being trained in games played with the unchanged card stats.

Other works have targeted understanding the intricacies of the design by learning from replays~\cite{bursztein2016legend} or at evaluating card uniqueness~\cite{janusz2018investigating}. An utility measurement to analyze card and deck usefulness in an effort to create a balance model has been proposed~\cite{jin2018proposed}. While this work treats cards individually, we are interested in the impact that changing the cards can have on prebuilt decks with preset strategies. 

Cards have been procedurally generated for Magic the Gathering from partial information~\cite{summerville2016mystical}. There is relation between card generation and card tuning and balancing, but our approach is not aimed at generating new content, but rather evaluating and changing the pre-existing cards.

\subsection{AI for Game Balancing}

Game balancing is a difficult task. As such, techniques using AI to facilitate or to try to automate the process have been introduced. Most work uses agents to collect data in order to evaluate the design and balance~\cite{de2017ai, silva2018exploring}.

Card games have been the focus of some of the most relevant work. Mahlmann et al. used an evolutionary algorithm to search a cardset in Dominion that would provide balanced gameplay~\cite{mahlmann2012evolving}. Jaffe et al. developed a tool that calculates balance metrics of the gameplay between restricted and standard agents, and applied such to an educational card game~\cite{jaffe2012evaluating}. Volz et al. used a multi-objective algorithm to create Top Trump decks, with win rate as one of the dimensions~\cite{volz2016demonstrating}. While these approaches are similar to the work we present, they are applied to games with lower complexity.

Other works have targeted understanding balance in a scope closer to ours. Krucher tried to automatically balance cards that were initially randomly generated~\cite{krucher2015algorithmically}. Our approach is not to make individual cards balanced, but investigate if we can make changes to the card attributes in order to balance win rate between different decks. Zook et al. investigated the impacts that changing a single card has on the design of Cardonomicon, a simplified version of Hearthstone~\cite{zook2018learning}. While their objective is to analyze how these changes reflect on player behavior, our work is targeted at understanding the level of impact changing cards can have on deck performance.

\section{Methodology and Definitions}

\subsection{Metagame and Match-ups}

 A \emph{Metagame} can be broadly defined as the set of factors outside the rules that affects the game experience. This can include the set of game objects (such as cards) that are collected by players, or less tangible factors such as heuristics on how to play a certain game situation (e.g. whether to prioritize killing Minions or dealing damage to the opposing Hero versus a specific deck) and even historical information about a certain player (e.g., upon facing the same player twice in a row in online match-making, there is a high likelihood they are playing the same deck as before).

This paper refers to a more narrow definition of metagame, focusing on the set of decks that players can build, their popularity, and expected win rate when each deck is paired against another. The metagame evolves over time as people discover new decks, and at any given time is simply called \emph{the meta} by the community. Sometimes decks that only differ by a few cards are referred to as a single deck or deck archetype. We avoid the distinction of decks and archetypes by assuming each deck is represented by a unique deck list.

A pair of decks and the expected win rate against each other is called a match-up, and the match-up between decks A and B is said to be favorable for deck A if deck A's win rate is above 50\%. Note that draws are possible, but rare in Hearthstone, so games that end in draws are ignored in our work. The match-up between two equal, or very similar decks is sometimes called \emph{the mirror}. We can also talk about a deck's win rate against the meta, which is the average win rate of all that deck's match-ups, weighted by the prevalence of the opposing deck in the metagame:
$$ W_i = \sum_{j=1}^N{W_{ij}P(j)}$$
where $W_i$ represents deck i's win rate versus the metagame, N is the total number of different decks, $W_{ij}$ is deck i's win rate versus deck j and $P(j)$ the probability of being paired with deck j in a match, also sometimes called deck j's prevalence or j's share of the metagame. Note that it includes mirror matches. In this paper, we consider $P(j) =\frac{1}{N}$ for all decks, meaning $W_i$ is simply the average win rate across all match-ups.

\subsection{Balance}

While determining whether or to what degree a game is balanced is a subject of debate by the community, one possible measure is that the metagame is more balanced as the win rates of all its match-ups approach 50\%. An alternative measure is that every deck's win rate versus the meta should be as close as possible to 50\% or that all decks take a similar share of the metagame. This definition allows us to formulate balance metrics based on the deck's pairwise match-ups, the deck's win rate against the meta, or a deck's prevalence. Because in most of our experiments we are dealing with metagames consisting of three distinct decks, we decided, for simplicity, to use pairwise match-up win rates as our metric. This will be defined in section~\ref{sec:exp1}.

While our experiments use this metric, this should not be taken as endorsement of this metric in the general case. A more nuanced view is that it is not the designer's job to perfectly balance all game objects, even if it were possible. Gutshera~\cite{gutschera} argues, in the context of balancing content for games such as Magic: The Gathering, that it can be desirable to have some game elements that are not viable (where viable means ``not strictly dominated''), because figuring out which objects are valuable can be an appealing aspect of gameplay. However, the viable objects are still expected to be balanced, while accepting that some game objects will be non-viable.

While in our experiments there are a fixed set of decks represented by a single deck list of thirty cards, the view above is especially relevant in more complex scenarios where this assumption doesn't hold. In this case, it might be important to consider not just how close each deck gets to perfect balance according to the chosen metric, but also how many reasonably balanced decks there are out of all possible decks, and how diverse these balanced decks are to each other.

We have so far assumed implicitly that the objects being balanced are decks and that we want decks to perform similarly to each other under the chosen metric. However, it is possible to balance around other objects. One could, for example, desire that a large number of different cards are playable and viable. One could also talk about the balance of strategies or heuristics, searching for metagames where a large number of strategies or heuristics are viable.

\subsection{Card Changes to Promote Game Balance}\label{changes}

While significant resources are expended during design to ensure balance, sometimes content creators misjudge the power level of a card when releasing a new set. Sometimes players find a hidden interaction between cards that leads to undesirable game states. Sometimes a metagame simply stagnates, with players converging on a consensus of what is the best deck, which halts innovation and leads to repetition. Regardless of the reason, sometimes action by the game's designers is necessary to restore balance, which is usually referred to as a balance change~\cite{ham2010rarity}.

In electronic games, a balance change typically consists of changing the attribute of one or more game objects to change their power level or to stop specific undesirable interactions between game objects. A change in card properties that makes a card more valuable is called a \emph{buff}, while one that decreases its value is called a \emph{nerf}. 

In physically distributed card games (such as Magic: The Gathering), changing a card's attribute is less practical, as it would lead to players owning an outdated copy of a card, which they might not even be aware has been changed. In such games, balance changes typically take the form of banning (prohibiting the use in sanctioned tournaments) or restricting (decreasing the number of copies that can be in a deck) cards.

In this paper, we consider balance changes defined by the increase or decrease in one or more of the following attributes of one or more cards:
\begin{itemize}
    \item \textbf{ Cost} is the amount of mana required to play each card.
    \item \textbf{Attack} is a property of minion and weapon cards and indicates how much damage each deals.
    \item \textbf{Health} is a property of minion cards and indicates how much damage the minion can sustain before it is destroyed.
    \item \textbf{Durability} is a property of weapon cards that indicates the number of times a weapon can be played.
\end{itemize}

An increase in mana cost or a decrease in attack or health is a nerf while a decrease in mana cost or an increase in attack or health is a buff.



Note that cards are fully defined by more than these properties and other properties of a card can be subject to balance changes. As an example, the neutral card, Patches the Pirate, was nerfed by losing the keyword \textit{Charge}, which allowed it to attack on the same turn it was summoned. Nevertheless, we choose to focus on mana cost, attack, health, and durability as they are shared by all cards of the appropriate type, have a simple representation and allow us to  express the majority of balance changes to cards in Hearthstone. Reference \cite{cardchanges:web19} lists all of the patches and updates of Hearthstone since those implemented in the alpha phase of development. Each patch catalogues all changes to cards, including changes that were motivated by balance.

\subsection{Magnitude of Balance Changes}

We define the absolute magnitude of a balance change as the sum of the absolute value of each change in attribute, weighted by that attribute's weight:

$$M = \sum{abs(C_i) * w_i}$$

Where $C_i$ is the change in an attribute and $w_i$ equals 2 if the attribute is mana cost and 1 otherwise. We use the weight $W$ because, in practice, changing a Minion's attack or health by 1 tends to have a smaller impact than changing its mana cost by 1. In fact, it is common that, for Minions that perform similar functions but differ by 1 mana, the more expensive minion will have +1/+1 in attack/health. For this reason, we consider a change in mana cost to have double the magnitude as a change in attack or health/Durability.

\section{Experimental Setup}

While several previous approaches explore the space of possible decks and creating competitive decks \cite{bhatt2018exploring, fontaine:gecco19,garcia2016evolutionary,garcia2018automated,chen2018q}, this paper investigates the impact of small changes to card properties on the deck's meta performance, where meta performance is measured as the average win rate against the other evolved decks.

Experiment 1 aims to maximize balance in the metagame through evolutionary search, where chromosomes are changes in properties of a card represented as integer vectors. Because it is important for players that their cards maintain some consistency, Experiment 2 searches for the minimal changes to balance the metagame. Experiment 2 replicates Experiment 1 with an added criteria of minimizing the magnitude of changes to the properties of the cards. Experiment 3 further isolates the deck space to target cards that need nerfs or buffs to minimize changes while simultaneously balancing the meta-game. The approach in this paper is to start from a set of competitive decks proposed by Fontaine et al.~\cite{fontaine:gecco19} and through evolution alter the properties of the cards to achieve balance in deck performance. 

\begin{table*}
\begin{tabular}{|p{1.8cm}|p{.7cm}|p{.7cm}|p{.85cm}|p{.85cm}|p{.95cm}|p{.95cm}|p{.7cm}|p{.7cm}|p{.85cm}|p{.85cm}|p{.95cm}|p{.95cm}|p{.7cm}|}
\hline
& Exp1 Hunter & Exp1 Hunter & Exp1 Paladin & Exp1 Paladin & Exp1 Warlock & Exp1 Warlock & Exp2 Hunter & Exp2 Hunter & Exp2 Paladin & Exp2 Paladin & Exp2 Warlock & Exp2 Warlock & Win rate vs. \\
& (Aggro) & (Control) & (Aggro) & (Control) & (Aggro) & (Control) & (Aggro) & (Control) & (Aggro) & (Control) &  (Aggro) & (Control) & Meta \\
\hline
Exp1:Hntr.(A) & 0.50 & 0.06 & 0.11 & 0.03 & 0.08 & 0.05 & 0.49 & 0.06 & 0.10 & 0.07 & 0.08 & 0.04 & 0.14 \\ \hline
Exp1:Hntr.(C) & 0.94 & 0.50 & 0.39 & 0.28 & 0.57 & 0.37 & 0.93 & 0.35 & 0.46 & 0.25 & 0.53 & 0.40 & 0.50 \\ \hline
Exp1:Pldn.(A) & 0.89 & 0.60 & 0.49 & 0.41 & 0.59 & 0.57 & 0.88 & 0.66 & 0.44 & 0.48 & 0.53 & 0.52 & 0.59 \\ \hline
Exp1:Pldn.(C) & 0.96 & 0.71 & 0.59 & 0.50 & 0.72 & 0.58 & 0.95 & 0.59 & 0.61 & 0.43 & 0.72 & 0.67 & 0.67 \\ \hline
Exp1:Wrlk.(A) & 0.91 & 0.42 & 0.42 & 0.28 & 0.50 & 0.36 & 0.88 & 0.36 & 0.37 & 0.29 & 0.38 & 0.38 & 0.46 \\ \hline
Exp1:Wrlk.(C) & 0.95 & 0.63 & 0.42 & 0.42 & 0.63 & 0.50 & 0.91 & 0.41 & 0.43 & 0.35 & 0.60 & 0.55 & 0.57 \\ \hline
Exp2:Hntr.(A) & 0.51 & 0.07 & 0.13 & 0.05 & 0.11 & 0.09 & 0.50 & 0.07 & 0.10 & 0.09 & 0.10 & 0.05 & 0.16 \\ \hline
Exp2:Hntr.(C) & 0.94 & 0.64 & 0.33 & 0.35 & 0.64 & 0.50 & 0.9 & 0.50 & 0.46 & 0.33 & 0.57 & 0.52 & 0.56 \\  \hline
Exp2:Pldn.(A) & 0.90 & 0.54 & 0.55 & 0.39 & 0.62 & 0.56 & 0.89 & 0.53 & 0.49 & 0.45 & 0.57 & 0.47 & 0.58 \\  \hline
Exp2:Pldn.(C) & 0.93 & 0.75 & 0.51 & 0.57 & 0.70 & 0.65 & 0.92 & 0.58 & 0.55 & 0.50 & 0.69 & 0.74 & 0.67 \\  \hline
Exp2:Wrlk(A) & 0.91 & 0.47 & 0.46 & 0.27 & 0.61 & 0.40 & 0.89 & 0.43 & 0.42 & 0.31 & 0.51 & 0.45 & 0.51 \\ \hline
Exp2:Wrlk(C) & 0.96 & 0.60 & 0.47 & 0.33 & 0.60 & 0.45 & 0.95 & 0.45 & 0.52 & 0.26 & 0.55 & 0.51 & 0.56 \\ \hline
\end{tabular}
\caption{Results from Pairwise Play of Twelve Competitive Decks 10,000 Games}
\label{table:initial10k}
\end{table*}

In Fontaine et al. \cite{fontaine:gecco19}, twelve decks were evolved for the hunter, paladin, and warlock classes playing an aggro and control strategy with two different opponent decks in evolution. These decks were evolved through a variant of the Multi-dimensional Archive of Phenotypic Elites (MAP-Elites) algorithm~\cite{mouret:nature15} called Map-Elites with Sliding Boundaries, and were first played against a pool of starter decks composed of basic cards, and then played against those evolved to beat these starter decks. The evolved decks had access to cards in the basic and classic sets.

We then decided to evaluate all these decks against each other. Performance is measured by pairwise match-ups, each played $10,000$ times as shown in table~\ref{table:initial10k}. Interestingly, gameplay heuristics have a significant impact on deck performance: all decks perform better when evolved with the Control heuristic. This is especially visible when comparing the performance of Hunter decks evolved with the Aggro and Control heuristic

For this paper, to reduce the space of decks, we selected one deck and heuristic combination for each hero is from these twelve competitive decks. These  will be our focus on for the remainder of the paper:Exp1: Hunter (Aggro), Exp2: Paladin (Control), and Exp2: Warlock (Control).

Table~\ref{table:smallmeta} shows each of these three deck's performance when considering only the match-ups against each other. From this point on, we will call the ``Small Meta" the meta-game consisting of these three decks or the decks evolved from them in the subsequent experiments. We will denote ``Original Meta" the meta-game defined by the original twelve decks.

\begin{table*}[]
\centering
\begin{tabular}{|l|l|l|l|l|}
\hline
\textbf{Deck Name}     & \textbf{Exp1: Hunter (Aggro)} & \textbf{Exp2: Paladin (Control)} & \textbf{Exp2: Warlock (Control)} & \textbf{Individual vs. Meta} \\ \hline
Exp1: Hunter (Aggro)    & 0.5                          & 0.0666                          & 0.0384                          & 0.2018666667                  \\ \hline
Exp2: Paladin (Control) & 0.9311                       & 0.5                             & 0.7381                          & 0.7227333333                  \\ \hline
Exp2: Warlock (Control) & 0.9622                       & 0.2648                          & 0.5                             & 0.578                         \\ \hline
\end{tabular}
\caption{The performance of each deck in the Small Meta}
\label{table:smallmeta}
\end{table*}

\subsection{Experiment 1: Meta Evolution}\label{sec:exp1}

The first experiment searches to maximize game balance by evolving properties of cards in the decks chosen to represent a low, high and medium level of competitiveness: Exp1: Hunter (Aggro), Exp2: Paladin (Control), Exp2: Warlock (Control). While there are 30 cards in each deck, there are 64 unique cards. Six are spells, 56 are minions, and two are weapons. Spell cards are adjusted through their mana cost properties, while minions and weapons are adjusted through their mana cost, attack, and health properties. While spell cards can potentially be altered through the magnitude of their effects, many effects are unique (e.g. Tirion Fordring's equip a 5/3 weapon, Archmage Antonidas's whenever you cast a spell, add a 'Fireball' spell to your hand). Future work aims to categorize the effects of these spells to evolve more comprehensive changes.

In these experiments, the population is composed of chromosomes that represent the changes to all card properties in the small meta. Here chromosomes are integer vectors of length 180 as there are three decks with sixty four unique cards. Six cards are spells with only one variable property, and 58 are weapon or minion cards with three variable properties. Therefore the length of a chromosome is calculated as 180 = 58 $\times$ 3 +  6 $\times$ 1. Each gene is an integer representation of changes to either mana cost, attack, or health properties in the range of $[-3,3].$ When Blizzard changes these card properties they often change in the range of $[-2,2]$, but in rare situations properties are changed by three. For these experiments, despite what a chromosome may indicate, card properties are bounded between $[0,10]$ .




The evolutionary parameters for the experiment are a population of size 100, crossover rate of 35\%, and mutation of 20\%. Crossover is two-point, and mutation changes each number with probability of 5\%. Parents are selected through tournament selection of size 3. Individuals are evaluated by first making changes to the cards in the decks and then playing 300 games in total, with 100 match-ups between Exp1: Hunter (Aggro) and Exp2: Paladin (Control), Exp2: Paladin (Control) and Exp2: Warlock (Control), and Exp1: Hunter (Aggro) and Exp2: Warlock (Control). Fitness is calculated as follows:
$$F = \sqrt{\frac{4}{3}\sum_{i<j}{(w_{ij}-0.5})^2}$$
where $w_{ij}$ is the win rate of deck $i$ vs. deck $j$, and $\frac{4}{3}$ is a constant for normalization so that $F \in [0,1]$. That is, fitness is calculated as the (normalized) Euclidean distance between the vector of candidate match-up win rates and an ideal vector where all entries equal $0.50$. Using this definition, our original small meta set has a fitness of $0.78$. While ultimately a good and playable meta decks would have some intransitivity \cite{bhatt2018exploring} (that is, deck A might beat deck B, which beats C, which in turn beats A), the goal for candidates is to achieve complete balance with a fitness of $0.00$ meaning that no changes are necessary to achieve a win rate of $0.50$.





\begin{figure}
    \centering
    \includegraphics[width=.4\textwidth]{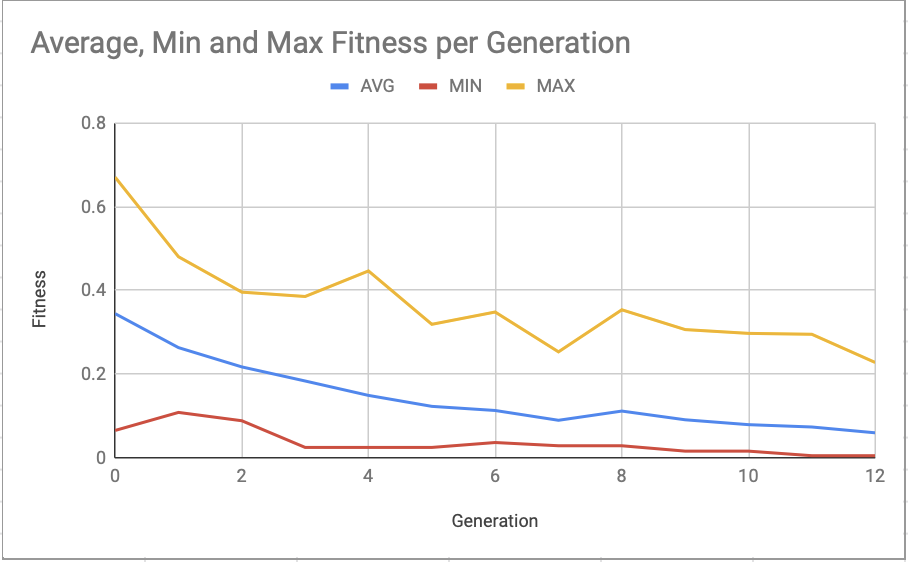}
    \caption{Average, Min and Max fitness over 12 generations. Baseline is the fitness of the original Small Meta, 0.78}
    \label{fig:exp1results}
\end{figure}

The results of experiment 1 are shown in figure~\ref{fig:exp1results}. The best individual across all generations had a fitness of $0.0062$, showing that the algorithm can find card change combinations leading to an almost perfectly balanced metagame. 

However upon examining this individual, the sum or magnitude of all changes to the 180 properties is 403, where the average individual in that generation had a magnitude of 402 (out of a maximum of 540). While these changes steer the meta to balance, the number of these changes would overwhelm a human designer and frustrate the game's players.

\subsection{Experiment 2: Multi-Objective Evolution} \label{sec:exp2}

\begin{figure}
    \centering
    \includegraphics[width=.4\textwidth]{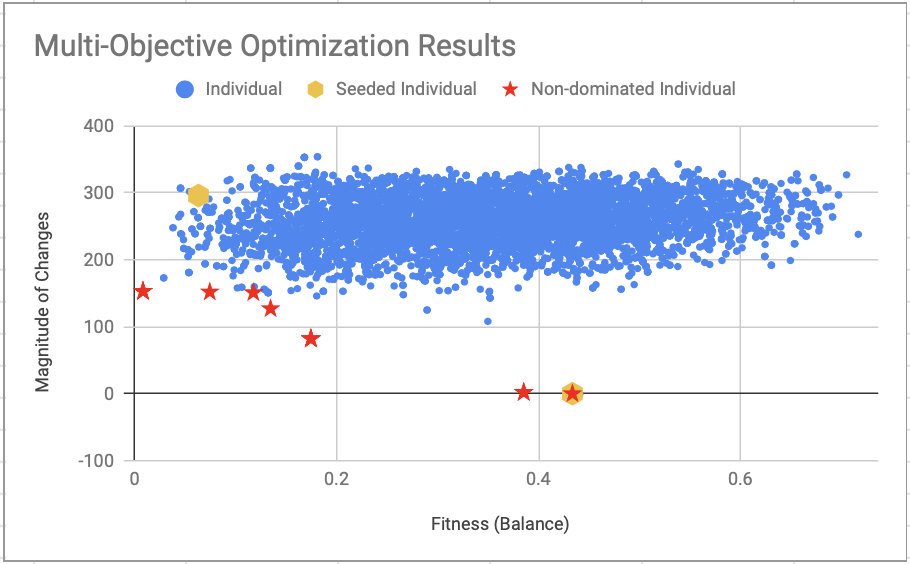}
    \caption{A scatter of points representing the individuals achieved through multi-objective optimization, with the Pareto Front (non-dominated individuals) highlighted in red. The seeded individuals (representing a metagame with no changes and the best metagame from Experiment 1) are highlighted in yellow. }
    \label{fig:paretofront}
\end{figure}


While experiment 1 found individuals that could steer the meta decks to a 50\% win rate, the distribution of the changes would be potentially cause large disruptions in current gameplay. Experiment 2 therefore aims to minimize changes while still achieving balance. Changes are measured by the sum of the values in the vector of an individual called its \emph{absolute magnitude}. 

To minimize the magnitude of changes, a multi-objective algorithm called NSGA2 \cite{deb2000fast} is run. An individual is dominated when there exists another that is strictly better (i.e. smaller values for changing card properties) across all card properties of the cards. NSGA2 identifies the set of non-dominated individuals when considering both the meta balance metric (defined by the distance of the pairwise win rates to $0.50$) and the magnitude of changes metric. Selection is performed by the NSGA2 algorithm, and  crossover and mutation follow the same rates and procedures of experiment 1 (i.e. two point crossover rate of 35\% and a mutation rate of 20\%. Each run was initialized with 98 randomly generated chromosomes, plus the best individual from experiment 1 (which had a balance metric of 0.0062 and a magnitude of changes of 402) and an individual representing the unchanging meta (which has a balance metric of 0.43 and, trivially, zero magnitude of changes).

An individual's  multi-objective fitness (MOF) is defined as
$$MOF_i = (F_i,M_i)$$
where $F$ is the previously defined meta balance metric and $M_i$ the magnitude of changes metric. Note that both $F$ and $M$ are metrics we are attempting to minimize. An individual $i$ is non-dominated if, for every other individual $j$ in the population, one of the conditions below hold:
\begin{itemize}
    \item $F_i < F_j$ or $M_i < M_j$ or ($F_i = F_j$ and $M_i = M_j$)
\end{itemize}

We executed two independent runs of this experiment, for 32 and 47 generations respectively.  Figure~\ref{fig:paretofront} shows the scatterplot of all individuals of both runs, with the set of individuals that were not dominated by any individual in each experiment (also called the Pareto Front) highlighted in red. Our individual with the lowest value for the meta balance dimension $F$ had a multi-objective fitness of $(0.008,154)$, and the individual with lowest value for the magnitude of changes dimension $M$ was the starting individual with no changes, who had a multi-objective fitness of  $(0.43,0)$. 

Note that our best individuals in Experiment 1 and Experiment 2 had about an equal balance score (0.006 in Experiment 1 and 0.008 in Experiment 2). Interestingly, this is not the same individual as the best individual from Experiment 1, which achieved a still low meta balance of 0.063 on this run. This is probably due to noise in the evaluation, which consist of playing 300 games per match-ups and comparing the win rates with the ideal 50\%. Nevertheless, we managed to achieve similar meta balance with less than half the magnitude of changes when compared to experiment 1 (154 versus 402). 



\subsection{Experiment 3: Heuristics for selecting cards to balance}\label{section:individualnerfs}

After showing in experiment 2 that a balanced metagame can be found with a smaller magnitude of changes than that in experiment 1, experiment 3 aims to predict which cards to target with a nerf or buff to achieve the best possible balance with the least possible changes.

For this experiment we define two auxiliary metrics that apply to cards in a deck, both listed by sites that aggregate Hearthstone replays (e.g. \url{http://HSReplay.net}). \emph{Win Rate when Played} (WRP) considers only games where at least one copy of the card is played. It is defined as the number of games won when at least one copy of the card is played divided by the total number of games where at least one copy of the card is played. It is a way to measure the positive ``impact" of a card once it is played. 


\emph{Win Rate when Drawn} (WRD) considers only games where at least one copy of the card is drawn (not necessarily played). It is defined as the number of games won where at least one copy of a card was drawn divided by the total number of games where at least one copy of that card was drawn. It is a way to measure the positive ``impact" of seeing a card in your hand, regardless of whether or not it was ever played over the course of the game. When compared to WRP, WRD tends to favor cheaper or more versatile cards, because cards that are expensive or situational tend to have a disproportionate impact in the game when successfully played, to compensate for the times where it cannot be played.

We define a card's \emph{Win Rate after Nerf} (WRN) as the deck's win rate when forced to play with a nerfed version of the card which costs one additional mana. This is meant to serve as a measure of a nerf's impact: the lower the WRN after nerfing a card, the more our targeted nerf was successful in balancing a dominating deck.

To compute WRN, we took every card in the Exp2 Paladin (Control) deck and, for each card ran a set of 10000 matches between Exp2 Paladin (Control) and the other 11 decks of the Original Meta, while increasing the nerfed card's mana cost by 1. Our hypothesis was that nerfing cards with higher WRP or WRD would have a bigger negative impact in the win rate of the deck. We also expected the impact to be greater for WRD, as WRD paints a more realistic picture of a card's power by taking into account situations where a card was drawn but could not be played.

For space considerations, we are unable to show the full relation of WRD, WRP and WRN for all the cards in the deck list. The card with highest WRD is Consecration, a four mana card that deals two damage to all enemy minions and the enemy hero, which serves as a check against an opponent who gets ahead on the board. The card wish highest WRP is Tirion Fordring, an expensive eight mana minion with high 6/6 stats, Taunt, Divine Shield (which prevents the first source of damage) and the ability to summon a powerful weapon when it dies. Interestingly, Consecration ranks third by WRP and Tirion Fordring ranks third by WRD. While they are both powerful cards, we suspect Consecration has WRD $>$ WRP because it is typically best played from an unfavorable position, when the opponent is ahead on board. Meanwhile, Tirion Fordring likely has WRP $>$ WRD due to it being high cost, making drawing it in the early turns a liability.

\begin{figure}
    \centering
    \subfigure[WRP versus WRN]
    {
        \includegraphics[scale=.4]{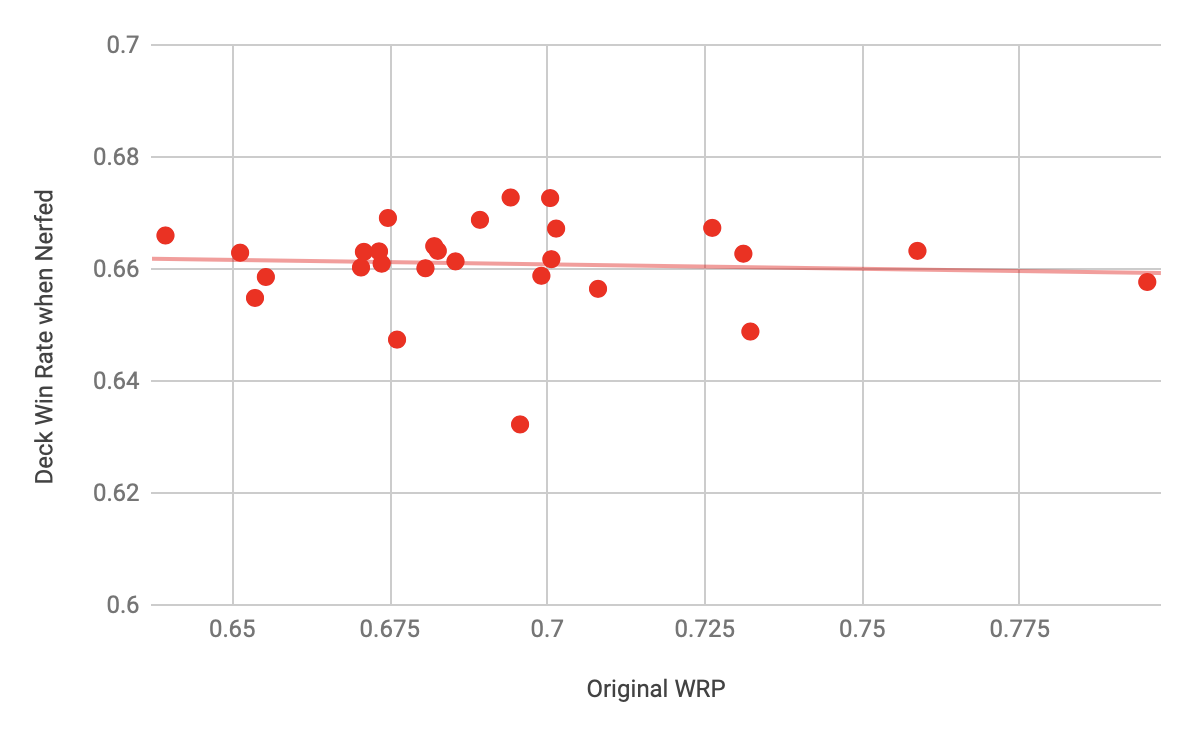}
        \label{fig:WRPversusWRN}
    }
    \subfigure[WRD versus WRN]
    {
        \includegraphics[scale=.4]{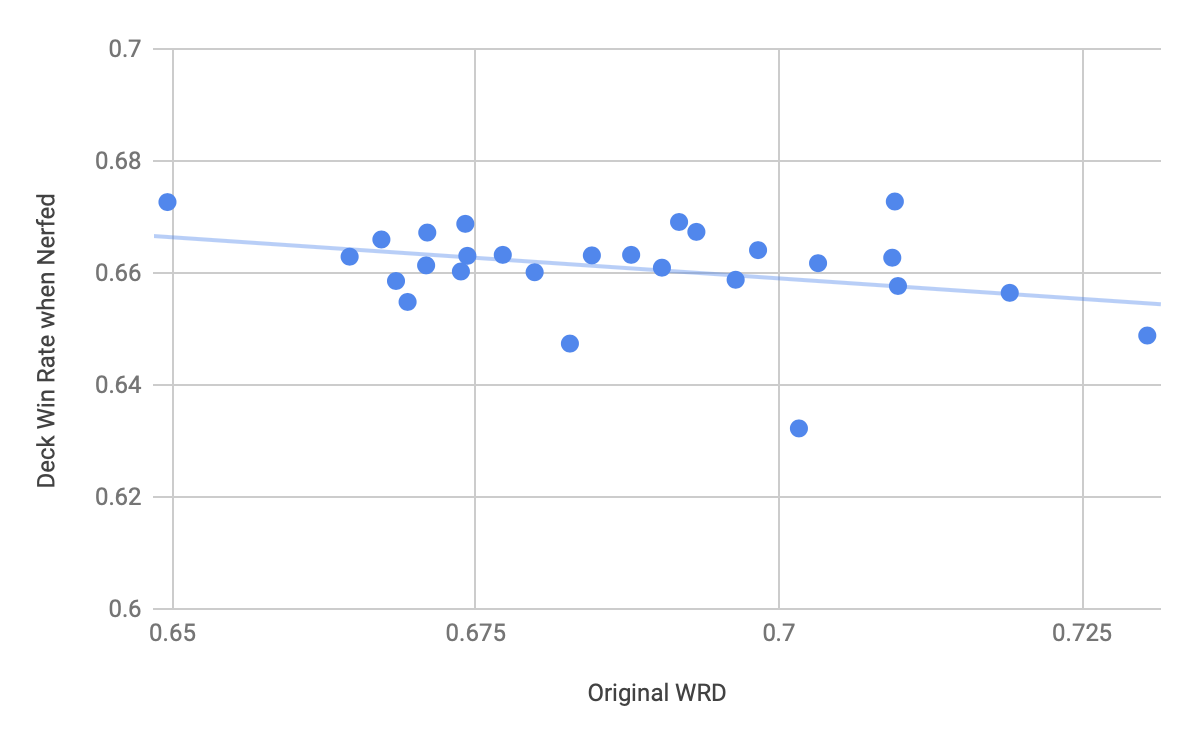}
        \label{fig:WRDversusWRN}
    }
    \caption{A scatterplot of WRP (a) and WRD (b) versus WRN}
    \label{fig:WRDandWRP}

\end{figure}

Figure~\ref{fig:WRDandWRP} shows the relationship, across all cards, of both WRD and WRP with the deck's win rate after the corresponding nerfs (WRN). Figure~\ref{fig:WRPversusWRN} shows a small negative correlation between WRP and WRN Figure~\ref{fig:WRDversusWRN} shows a slightly bigger negative correlation between WRD and WRN. A negative correlation means the deck's win rate tends to fall more sharply when a card with high WRP or WRD is nerfed, which indicates that nerfs to those cards are more effective at balancing a deck that is too dominant. While we make no claims of statistical significance of these findings, the data seems to support our initial assumptions.

\section{Conclusion and Future Work}

This paper proposes a method for representing balance changes in a card game such as Hearthstone and seeks to minimize the number of changes through evolutionary search while maximizing metagame balance (section~\ref{sec:exp1}). While this approach successfully tuned cards to create a balanced metagame, the tuning process in experiment 1 suggested a disruptive magnitude of changes. 
To compensate, in experiment 2 card changes are evolved through multi-objective evolution with the NSGA2 algorithm
to find balance changes that simultaneously optimize for balanced metagames and minimized magnitudes of these changes.

The experiments in section~\ref{section:individualnerfs} gives us some indication that win rate when played (WRP) and especially win rate when drawn (WRD) are metrics that can help choose which cards to nerf, which in the future we expect to use as a heuristic to guide the search for a set of maximally-balancing, minimally-disruptive balance changes.

However all experiments assume that a player's deck remains the same after balance changes are implemented. However, players often respond to balance changes by altering their decks. Future work will explore deck composition in response to a balance changes by evolving decklists after applying the balance changes compared to the original metagame.

Finally, this paper defines a balanced metagame as one where all decks have close to 50\% win rate against all other decks. However, future work will explore options like optimizing a deck's \emph{average} win rate against all decks, but permit deck intransitivity \cite{de2018generatingprefloppoker,de2018generatingfullpoker} by allowing differences in a deck's efficacy against \emph{particular} other decks. We could also incorporate some metric of deck diversity (such as the number of shared cards) and require the metagame to have a large number of decks that are not only balanced with regard to their win rate, but also highly diverse. One final approach would be to formulate the problem of choosing a deck from the metagame as a two-player game in extensive form, and compute the Mixed Nash Equilibrium~\cite{nash1951non} for this game, requiring that, in the equilibrium, a large number of decks are non-dominated and have a reasonable chance of being picked.


\bibliographystyle{IEEEtran}
\bibliography{bibfile}

\begin{thebibliography}{10}
\providecommand{\url}[1]{#1}
\csname url@samestyle\endcsname
\providecommand{\newblock}{\relax}
\providecommand{\bibinfo}[2]{#2}
\providecommand{\BIBentrySTDinterwordspacing}{\spaceskip=0pt\relax}
\providecommand{\BIBentryALTinterwordstretchfactor}{4}
\providecommand{\BIBentryALTinterwordspacing}{\spaceskip=\fontdimen2\font plus
\BIBentryALTinterwordstretchfactor\fontdimen3\font minus
  \fontdimen4\font\relax}
\providecommand{\BIBforeignlanguage}[2]{{%
\expandafter\ifx\csname l@#1\endcsname\relax
\typeout{** WARNING: IEEEtran.bst: No hyphenation pattern has been}%
\typeout{** loaded for the language `#1'. Using the pattern for}%
\typeout{** the default language instead.}%
\else
\language=\csname l@#1\endcsname
\fi
#2}}
\providecommand{\BIBdecl}{\relax}
\BIBdecl

\bibitem{da2018hearthbot}
A.~R. da~Silva and L.~F.~W. Goes, ``Hearthbot: An autonomous agent based on
  fuzzy art adaptive neural networks for the digital collectible card
  gamehearthstone,'' \emph{IEEE Transactions on Games}, vol.~10, no.~2, pp.
  170--181, 2018.

\bibitem{dockhorn2018predicting}
A.~Dockhorn, M.~Frick, {\"U}.~Akkaya, and R.~Kruse, ``Predicting opponent moves
  for improving hearthstone ai,'' in \emph{International Conference on
  Information Processing and Management of Uncertainty in Knowledge-Based
  Systems}.\hskip 1em plus 0.5em minus 0.4em\relax Springer, 2018, pp.
  621--632.

\bibitem{grad2017helping}
{\L}.~Grad, ``Helping ai to play hearthstone using neural networks,'' in
  \emph{2017 federated conference on computer science and information systems
  (FedCSIS)}.\hskip 1em plus 0.5em minus 0.4em\relax IEEE, 2017, pp. 131--134.

\bibitem{santos2017monte}
A.~Santos, P.~A. Santos, and F.~S. Melo, ``Monte carlo tree search experiments
  in hearthstone,'' in \emph{Computational Intelligence and Games (CIG), 2017
  IEEE Conference on}.\hskip 1em plus 0.5em minus 0.4em\relax IEEE, 2017, pp.
  272--279.

\bibitem{stiegler2017symbolic}
A.~Stiegler, K.~Dahal, J.~Maucher, and D.~Livingstone, ``Symbolic reasoning for
  hearthstone,'' \emph{IEEE Transactions on Computational Intelligence and AI
  in Games}, 2017.

\bibitem{swiechowski2018improving}
M.~{\'S}wiechowski, T.~Tajmajer, and A.~Janusz, ``Improving hearthstone ai by
  combining mcts and supervised learning algorithms,'' in \emph{2018 IEEE
  Conference on Computational Intelligence and Games (CIG)}.\hskip 1em plus
  0.5em minus 0.4em\relax IEEE, 2018, pp. 1--8.

\bibitem{zhang2017improving}
S.~Zhang and M.~Buro, ``Improving hearthstone ai by learning high-level rollout
  policies and bucketing chance node events,'' in \emph{Computational
  Intelligence and Games (CIG), 2017 IEEE Conference on}.\hskip 1em plus 0.5em
  minus 0.4em\relax IEEE, 2017, pp. 309--316.

\bibitem{bhatt2018exploring}
A.~Bhatt, S.~Lee, F.~de~Mesentier~Silva, C.~W. Watson, J.~Togelius, and A.~K.
  Hoover, ``Exploring the hearthstone deck space,'' in \emph{Proceedings of the
  13th International Conference on the Foundations of Digital Games}.\hskip 1em
  plus 0.5em minus 0.4em\relax ACM, 2018, p.~18.

\bibitem{fontaine:gecco19}
M.~F.~S. Lee, L.~B. S.~F. de~Mesentier~Silva, and J.~T. A.~K. Hoover, ``Mapping
  hearthstone deck spaces with map-elites with sliding boundaries,'' in
  \emph{To be published at Proceedings of The Genetic and Evolutionary
  Computation Conference}.\hskip 1em plus 0.5em minus 0.4em\relax ACM, 2019,
  p.~8.

\bibitem{garcia2016evolutionary}
P.~Garc{\'\i}a-S{\'a}nchez, A.~Tonda, G.~Squillero, A.~Mora, and J.~J. Merelo,
  ``Evolutionary deckbuilding in hearthstone,'' in \emph{Computational
  Intelligence and Games (CIG), 2016 IEEE Conference on}.\hskip 1em plus 0.5em
  minus 0.4em\relax IEEE, 2016, pp. 1--8.

\bibitem{garcia2018automated}
P.~Garc{\'\i}a-S{\'a}nchez, A.~Tonda, A.~M. Mora, G.~Squillero, and J.~J.
  Merelo, ``Automated playtesting in collectible card games using evolutionary
  algorithms: A case study in hearthstone,'' \emph{Knowledge-Based Systems},
  vol. 153, pp. 133--146, 2018.

\bibitem{mouret2015illuminating}
J.-B. Mouret and J.~Clune, ``Illuminating search spaces by mapping elites,''
  \emph{arXiv preprint arXiv:1504.04909}, 2015.

\bibitem{chen2018q}
Z.~Chen, C.~Amato, T.-H.~D. Nguyen, S.~Cooper, Y.~Sun, and M.~S. El-Nasr,
  ``Q-deckrec: A fast deck recommendation system for collectible card games,''
  in \emph{2018 IEEE Conference on Computational Intelligence and Games
  (CIG)}.\hskip 1em plus 0.5em minus 0.4em\relax IEEE, 2018, pp. 1--8.

\bibitem{stiegler2016hearthstone}
A.~Stiegler, C.~Messerschmidt, J.~Maucher, and K.~Dahal, ``Hearthstone
  deck-construction with a utility system,'' in \emph{Software, Knowledge,
  Information Management \& Applications (SKIMA), 2016 10th International
  Conference on}.\hskip 1em plus 0.5em minus 0.4em\relax IEEE, 2016, pp.
  21--28.

\bibitem{bjorke2017deckbuilding}
S.~J. Bj{\o}rke and K.~A. Fludal, ``Deckbuilding in magic: The gathering using
  a genetic algorithm,'' Master's thesis, Norwegian University of Science and
  Technology (NTNU), 2017.

\bibitem{jakubik2017evaluation}
J.~Jakubik, ``Evaluation of hearthstone game states with neural networks and
  sparse autoencoding,'' in \emph{Computer Science and Information Systems
  (FedCSIS), 2017 Federated Conference on}.\hskip 1em plus 0.5em minus
  0.4em\relax IEEE, 2017, pp. 135--138.

\bibitem{janusz2018toward}
A.~Janusz, T.~Tajmajer, M.~{\'S}wiechowski, {\L}.~Grad, J.~Puczniewski, and
  D.~{\'S}lezak, ``Toward an intelligent hs deck advisor: Lessons learned from
  aaia’18 data mining competition,'' in \emph{2018 Federated Conference on
  Computer Science and Information Systems (FedCSIS)}.\hskip 1em plus 0.5em
  minus 0.4em\relax IEEE, 2018, pp. 189--192.

\bibitem{bursztein2016legend}
E.~Bursztein, ``I am a legend: Hacking hearthstone using statistical learning
  methods,'' in \emph{Computational Intelligence and Games (CIG), 2016 IEEE
  Conference on}.\hskip 1em plus 0.5em minus 0.4em\relax IEEE, 2016, pp. 1--8.

\bibitem{janusz2018investigating}
A.~Janusz and D.~Slezak, ``Investigating similarity between hearthstone cards:
  Text embeddings and interchangeability approaches,'' in \emph{2018 IEEE
  International Conference on Systems, Man, and Cybernetics (SMC)}.\hskip 1em
  plus 0.5em minus 0.4em\relax IEEE, 2018, pp. 3421--3426.

\bibitem{jin2018proposed}
Y.~Jin, ``Proposed balance model for card deck measurement in hearthstone,''
  \emph{The Computer Games Journal}, pp. 1--16, 2018.

\bibitem{summerville2016mystical}
A.~J. Summerville and M.~Mateas, ``Mystical tutor: A magic: The gathering
  design assistant via denoising sequence-to-sequence learning,'' in
  \emph{Twelfth Artificial Intelligence and Interactive Digital Entertainment
  Conference}, 2016.

\bibitem{de2017ai}
F.~de~Mesentier~Silva, S.~Lee, J.~Togelius, and A.~Nealen, ``Ai-based
  playtesting of contemporary board games,'' in \emph{Proceedings of the 12th
  International Conference on the Foundations of Digital Games}.\hskip 1em plus
  0.5em minus 0.4em\relax ACM, 2017, p.~13.

\bibitem{silva2018exploring}
F.~D.~M. Silva, I.~Borovikov, J.~Kolen, N.~Aghdaie, and K.~Zaman, ``Exploring
  gameplay with ai agents,'' in \emph{Fourteenth Artificial Intelligence and
  Interactive Digital Entertainment Conference}, 2018.

\bibitem{mahlmann2012evolving}
T.~Mahlmann, J.~Togelius, and G.~N. Yannakakis, ``Evolving card sets towards
  balancing dominion,'' in \emph{2012 IEEE Congress on Evolutionary
  Computation}.\hskip 1em plus 0.5em minus 0.4em\relax IEEE, 2012, pp. 1--8.

\bibitem{jaffe2012evaluating}
A.~Jaffe, A.~Miller, E.~Andersen, Y.-E. Liu, A.~Karlin, and Z.~Popovic,
  ``Evaluating competitive game balance with restricted play,'' in
  \emph{Proceedings of the Eighth Artificial Intelligence and Interactive
  Digital Entertainment International Conference (AIIDE 2012)}, 2012.

\bibitem{volz2016demonstrating}
V.~Volz, G.~Rudolph, and B.~Naujoks, ``Demonstrating the feasibility of
  automatic game balancing,'' in \emph{Proceedings of the 2016 on Genetic and
  Evolutionary Computation Conference}.\hskip 1em plus 0.5em minus 0.4em\relax
  ACM, 2016, pp. 269--276.

\bibitem{krucher2015algorithmically}
J.~Krucher, ``Algorithmically balancing a collectible card game,''
  \emph{Bachelor’s Thesis. ETH Zurich}, 2015.

\bibitem{zook2018learning}
A.~Zook and M.~Riedl, ``Learning how design choices impact gameplay behavior,''
  \emph{IEEE Transactions on Games}, 2018.

\bibitem{gutschera}
K.~R. Gutschera, ``Magic lessons: Designing and balancing game objects,''
  \url{http://twvideo01.ubm-us.net/o1/vault/gdc07/slides/S3709i2.pdf},
  accessed: 2019-03-24.

\bibitem{ham2010rarity}
E.~Ham, ``Rarity and power: balance in collectible object games,'' \emph{The
  International Journal of Computer Game Research}, vol.~10, no.~1, 2010.

\bibitem{cardchanges:web19}
``Card changes,'' \url{https://hearthstone.gamepedia.com/Card\_changes},
  accessed: 2019-03-24.

\bibitem{mouret:nature15}
J.-B. Mouret and J.~Clune, ``Illuminating search spaces by mapping elites,''
  \emph{arXiv preprint arXiv:1504.04909}, 2015.

\bibitem{deb2000fast}
K.~Deb, S.~Agrawal, A.~Pratap, and T.~Meyarivan, ``A fast elitist non-dominated
  sorting genetic algorithm for multi-objective optimization: Nsga-ii,'' in
  \emph{International conference on parallel problem solving from
  nature}.\hskip 1em plus 0.5em minus 0.4em\relax Springer, 2000, pp. 849--858.

\bibitem{de2018generatingprefloppoker}
F.~de~Mesentier~Silva, J.~Togelius, F.~Lantz, and A.~Nealen, ``Generating
  beginner heuristics for simple texas hold’em,'' 2018.

\bibitem{de2018generatingfullpoker}
------, ``Generating novice heuristics for post-flop poker,'' in \emph{2018
  IEEE Conference on Computational Intelligence and Games (CIG)}.\hskip 1em
  plus 0.5em minus 0.4em\relax IEEE, 2018, pp. 1--8.

\bibitem{nash1951non}
J.~Nash, ``Non-cooperative games,'' \emph{Annals of mathematics}, pp. 286--295,
  1951.

\end{thebibliography}

\end{document}